\documentclass[letterpaper]{article} % DO NOT CHANGE THIS
\usepackage{aaai24}  % DO NOT CHANGE THIS
\usepackage{times}  % DO NOT CHANGE THIS
\usepackage{helvet}  % DO NOT CHANGE THIS
\usepackage{courier}  % DO NOT CHANGE THIS
\usepackage[hyphens]{url}  % DO NOT CHANGE THIS
\usepackage{graphicx} % DO NOT CHANGE THIS
\def\UrlFont{\rm}  % DO NOT CHANGE THIS
\usepackage{natbib}  % DO NOT CHANGE THIS AND DO NOT ADD ANY OPTIONS TO IT
\usepackage{caption} % DO NOT CHANGE THIS AND DO NOT ADD ANY OPTIONS TO IT
\usepackage{algorithm}
\usepackage{algorithmic}
\usepackage{amsmath}
\usepackage{newfloat}
\usepackage{listings}
\DeclareCaptionStyle{ruled}{labelfont=normalfont,labelsep=colon,strut=off} % DO NOT CHANGE THIS
\floatstyle{ruled}
\newfloat{listing}{tb}{lst}{}
\floatname{listing}{Listing}
\title{Pattern-Based Time-Series Risk Scoring for Anomaly Detection and Alert Filtering - A Predictive Maintenance Case Study}%
\author{Elad Liebman \\ SparkCognition Research \\ Austin, TX, USA} 

\begin{document}

\maketitle
\begin{abstract}
Fault detection is a key challenge in the management of complex systems. In the context of SparkCognition's efforts towards predictive maintenance in large scale industrial systems, this problem is often framed in terms of anomaly detection - identifying patterns of behavior in the data which deviate from normal. Patterns of normal behavior aren't captured simply in the coarse statistics of measured signals. Rather, the multivariate sequential pattern itself can be indicative of normal vs. abnormal behavior. For this reason, normal behavior modeling that relies on snapshots of the data without taking into account temporal relationships as they evolve would be lacking. However, common strategies for dealing with temporal dependence, such as Recurrent Neural Networks or attention mechanisms are oftentimes computationally expensive and difficult to train. In this paper, we propose a fast and efficient approach to anomaly detection and alert filtering based on sequential pattern similarities. In our empirical analysis section, we show how this approach can be leveraged for a variety of purposes involving anomaly detection on a large scale real-world industrial system. Subsequently, we test our approach on a publicly-available dataset in order to establish its general applicability and robustness compared to a state-of-the-art baseline. We also demonstrate an efficient way of optimizing the framework based on an alert recall objective function.
\end{abstract}

\maketitle

\section{Introduction and Problem Definition}

Fault detection is a key challenge in the management of complex systems. In the context of SparkCognition's predictive maintenance modeling this problem is often framed as \emph{anomaly detection} - identifying patterns of behavior in the data which deviate from normal. Alas, not every deviation is necessarily indicative of an actual malfunction, and therefore we must differentiate between two distinct terms:

\begin{itemize}

\item \emph{events} - actual recorded cases of inappropriate system behavior

\item \emph{alerts} - sequences in time in which the maintenance model detects abnormal behavior

\end{itemize}

One of the core objectives of a predictive maintenance framework is to alert either right before or as early as possible after an abnormal event takes place, and sustain the alert until the event is concluded. 

In this study we propose a lightweight time-series framework for alert management and potentially as a second tier of defense for event detection. We further show its usefulness on large-scale, highly complex datasets capturing physical properties of real-world systems.

\section{Related Work}

Due to its many applications, anomaly detection has attracted considerable attention in the past two decades ~\cite{chandola2009anomaly, blazquez2020review}. To mention some relevant examples from the past few years, 
Doan et al. employed ensemble switching and grid-based clustering for anomaly detection in pedestrian distribution data~\cite{doan2015profiling}. Su et al. \ employed stochastic RNNs towards anomaly detection in server machine and aerospace datasets~\cite{su2019robust}. Hundman et al. \ utilized LSTMs and nonparametric dynamic error thresholding to detect anomalies in spacecraft. Ahmad et al.\ offer a different approach based on Hierarchical Temporal Memory~\cite{ahmad2017unsupervised}.  Qiu et al.\ study a novel architecture which combines LSTMs, CNNs and Variational Inference Autoencoders (VAE) and show its effectiveness in both the A1benchmark and A2benchmark datasets as well as some real-world production datasets~\cite{qiu2019kpi}. Another family of approaches involve Temporal Convolutional Networks (TCNs). For example,  He and Zhao leverage TCNs for anomaly detection in EKG,  space shuttle and gesture datasets~\cite{he2019temporal}.Quite recently, Kong et al.\ introduced a combination of attention mechanisms, generative methods and LSTMs towards anomaly detection in industrial datasets~\cite{kong2021integrated}.

From a more pattern-based perspective, Yeh et al.\ propose an all-pairs FFT-based similarity matrix profile approac which also has applicability toward shapelet identification~\cite{yeh2016matrix}, whereas Farahani et al.\ propose a probabilistic approach based on clustering the overlapping subsequences of the training dataset and analyzing their orders by Markov chains and show its applicability in medical, utility and transport demand datasets \cite{farahani2019time}.

While this list of references is somewhat representative of the breadth of approaches proposed in order to deal with anomaly detection in complex environments, it is far from comprehensive. For a more detailed review please see Blazquez-Garcia et al.~\cite{blazquez2020review}.

\subsection{Limitations of Current Approaches}
\label{sec:lim}
While many of the approaches discussed in the section above show promise on certain applications, the real-world datasets SparkCognition deals with on a daily basis are challenging even by current research standards. The difficulties with these datasets stem from several key issues: 

\begin{itemize}

\item \emph{Size} - even by industry standards, many of the datasets our models are expected to ingest are very large, containing many millions of records and spanning dozens of thousands of parameters. These limitations render most of the leading approaches in the literature computationally intractable. This limitation is particularly true for graphical models,  deep neural network and traditional signal processing approaches. In practice most baseline approaches we attempted failed for computational and practical reasons off the bat, being unable to handle extremely large quantities of multivariate data.

\item \emph{Structural Sparsity} - beyond data size rendering many approaches computationally infeasible, it also induces a \emph{highly sparse and complex state space}. As Heraclitus once said,  ``No man ever steps in the same river twice, for it's not the same river and he's not the same man'', this is often true for the type of datasets we encounter - you never see the same datapoint twice, for it is not the same datapoint and, especially for adaptive models, it is not the same model either, typically.  What this limitation implies in practice is that it is very hard to utilize approaches that rely on successfully identifying common patterns or modes of operation. In practice, clustering-based approach tend to either generate clumps of incoherent clusters or many sparsely-represented clusters, rarely finding any convincing sweet spot.

\item \emph{Event/Labeling Sparsity} - even by industry standards the datasets we work with tend to include barely a handful of labeled anomalous events. For a given system, it is not uncommon to receive datasets spanning years but which only contain 3-4 labeled events which are considered by the annotator to be actual reportable anomalies.  This sparsity in ground truth, vetted events is a huge challenge for any supervised learning approach, but if one takes an unsupervised (or weakly supervised) approach, the fact that you only have very few actual confirmed anomalies means evaluating your algorithm is nontrivial.

\item \emph{Explainability} - While explainability is certainly not a novel concern in machine learning, nor is it new in the context of anomaly detection, we have found that it is under studied compared to the real-world concerns involved in predictive maintenance for large scale industrial systems. While a timely warning describing anomalous behavior may save a complex system from disaster, if it is not coupled with compelling justification as for \emph{why} the model would generate such a prediction, it is less likely to draw actionable response from users.

\end{itemize}

In the next section we discuss our general approach and how it is designed to mitigate the concerns discussed above.

\section{Methods}
\label{sec:method}

As noted in some previous work~\cite{farahani2019time}, the characteristics of normal behavior aren't always straightforwardly captured in the coarse statistics of measured signals. Rather, the multivariate \emph{sequential pattern} itself can be indicative of normal vs. abnormal behavior. However, as discussed in Section \ref{sec:lim},  many cutting edge strategies for anomaly detection, ranging from graphical models and signal processing to LSTMs, TCNs and attention are computationally intractable and difficult to train in our case. Nonetheless, normal behavior modeling that relies on snapshots of the data without taking into account temporal relationships as they evolve would be lacking. Instead, we focus on a fast and efficient \emph{pattern-based} approach, based on breaking the stream of data into multivariate overlapping windows, and identifying sequential similarities based on selected, computationally efficient similarity measures. 

The core idea is that while it is infeasible to perform any many-to-many comparisons between patterns in complex data, taking a \emph{smoothed multivariate subsequence} and comparing it to ones preceding can be done efficiently. Furthermore, we can reduce the number of comparisons by focusing on features of interest, and by limiting our historical comparisons based on data drift and computational resources.If you consider how different a new subsequence is \emph{from the previous subsequence most similar to it}, that difference can serve as an anomaly score. We further show in Section \ref{sec:alsim} how similarities between windows can be further leveraged for annotation propagation and improved event recall.

\begin{figure}[!htb]
\centering
\includegraphics[width=\columnwidth, height=240pt]{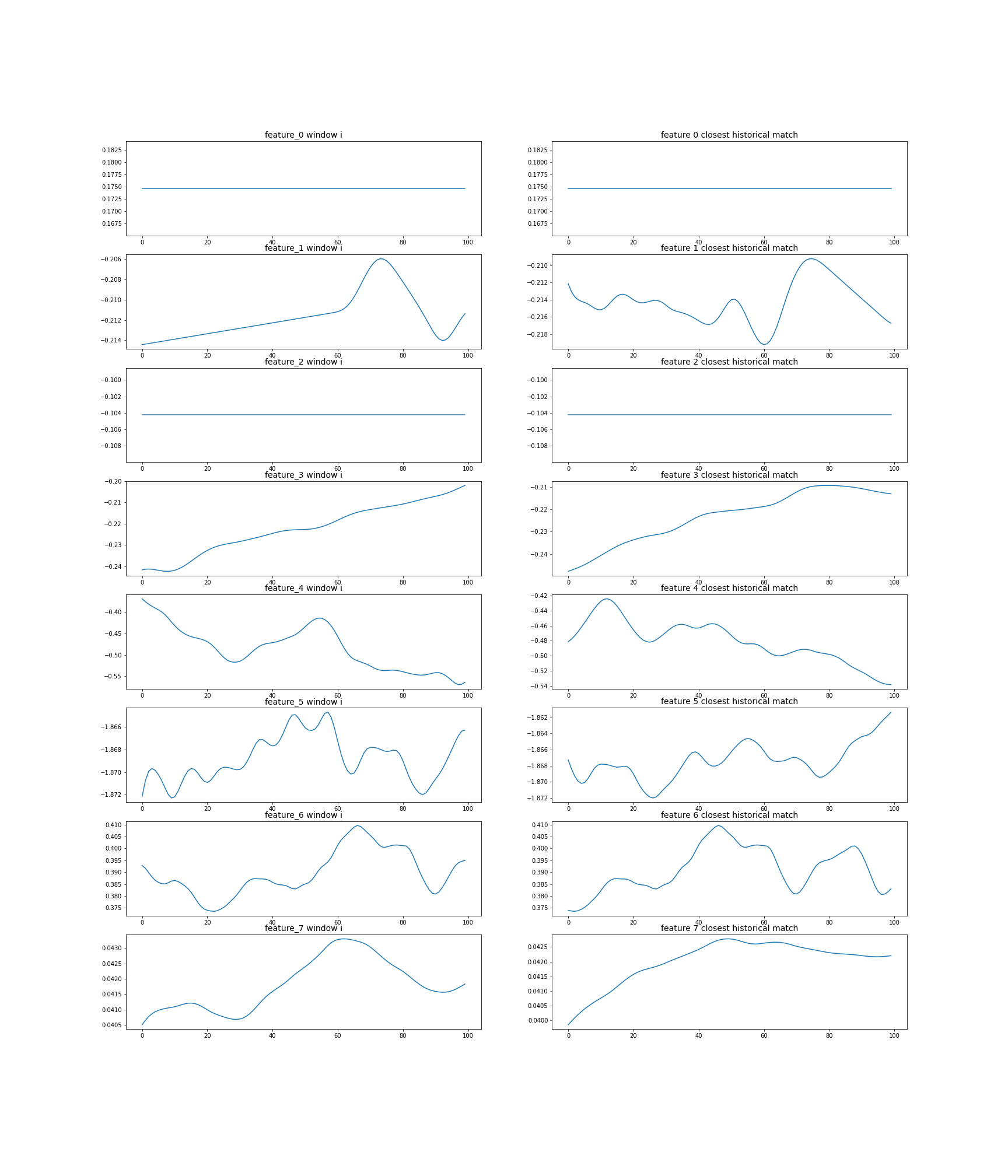}%{Figure6.png}
\caption{
A representation of comparing two similar windows based on smoothed top (i.e. most informative) features. On the left is the target window, and on the right the closest matching window preceding it based on our proposed cross-correlation distance. Each row represents a different feature of the data (feature types and characteristics have been anonymized at the customer’s behest but they represent physical properties of a system such as voltage, temperature and pressure). The X axis represents time (100 seconds).
}
%A representation of comparing two similar windows based on smoothed top features. On the left is the target window, and on the right the closest matching window preceding it based on cross-correlation distance.}
\label{fig1}

\end{figure}

\subsection{Core Algorithm}
\label{sec:core}
The steps of this approach are as follows:

\begin{itemize}

\item At the first step, we smooth each feature to better identify trends within the data. This is accomplished using a Hanning smoothing approach~\cite{jekeli1981alternative}. This step serves to mitigate the sensitivity of window comparisons to transient noise which obfuscates the underlying patterns.

\item At the second step, the smoothed multivariate data stream is sliced into overlapping subsequences, or windows. For computational purposes, the extent of overlap is a controllable parameter which can be tuned based on computational requirements and data structure. 

Furthermore, rather than generating overlapping windows for the entire dataset, one can limit window generation based on human annotation (experts labeling the data stream for suspected regions), or the output of another model which provides some risk-scoring signal (another anomaly detection model, essentially), meaning our approach is easily composable with other anomaly detection methods.

\item Given a new window $w_i$, we compare it to all relevant previous prespecified windows $W_{j\neq i} = w_{i-t}, w_{i-t+1}, \ldots, w_{i-1}$.

\item Let $w_1, w_2$ be two multivariate windows we wish to compare. The choice of distance measure is a parameter and the general framework is robust to changes in metric. However, in our implementation, we focused on the following distance measures:
\begin{itemize}
\item Euclidean distance between the multivariate alignments (this option assumes the data is normalized), $||w_1 - w_2||$.  If one window is smaller than the other then it is slid over the larger one and either the average or the minimum are taken.
\item Dynamic time warping \cite{muller2007dynamic}
\item the normalized negative of the maximum of the cross-correlation, $(w_1 \star w_2)(\tau)/||w_1 \cdot w_2||)$. 
\end{itemize}

Once again, we note that if the lengths of $w_1, w_2$ are not equal the shorter one is gradually offset across the other one and the distance is computed at different offsets. While it can be a single timestep (of length depending on the sampling rate of the data), the sliding offset for this procedure is parameterized and for speed-up purposes it is often useful to set it at a higher value than $1$.  To save computation, one can limit these comparisons to the $k$ most relevant shared features between the two windows. Moreover, for speed-up purposes, these distance computations can be done for each dimension separately and then aggregated (if the lengths are not equal, they are aggregated for each timestep separately). 

\item Lastly, to get a baseline risk score for each window, $w_i$ retrieve its distance from \emph{the closest relevant window preceding it} $\textbf{MIN}_{w_j \in W_{j\neq i} }d(w_i,w_j)$.  A naive approach for alerting would involve thresholding this time-series deviation score.

\end{itemize}

A pseudocode summary of this time-series deviation scoring algorithm is presented in Algorithm \ref{alg1}.
\begin{algorithm}
\caption{Time deviation scoring by contour matching}
\label{alg1}
\begin{algorithmic}[1]
 \REQUIRE $k$-dimensional multivariate time series data $\mathcal{D}$, smoothing parameters $\theta$, windows $\tau_1, \tau_2, \ldots, \tau_M$
 \STATE smooth data:
 \FOR{$i = 1$ to $k$}
     \STATE $\mathcal{D}_k \gets \textit{smooth}_\theta(\mathcal{D}_k)$
 \ENDFOR
 \STATE compute windows:
 \FOR{$j = 1$ to $M$}
     \STATE $\mathbf{W_j}(\mathbf{D}) \gets \text{compute windows by } \tau_j$
     \FOR{each $w_l \in \mathbf{W_j}(\mathbf{D})$}
         \STATE compute distance from $w_l$ to $w_1, \ldots, w_{l-1}$:
         \STATE $W_{r \neq l} = \textit{d}_r(w_r, w_l) \quad \forall r < l$
         \STATE compute time deviation score:
         \STATE $s(w_l) = \min_{r < l}(\textit{d}_r(w_r, w_l) \in W_{r \neq l})$
     \ENDFOR
 \ENDFOR
\end{algorithmic}
\end{algorithm}

A visualization of this type of matching and comparison is shown in Figure \ref{fig1}.

\subsection{Alert Similarity, Majority Voting, and Filtering}
\label{sec:alsim}
The main insight underlying our approach is that given a pair of alerts, one can compare them as described in Section \ref{sec:method}.Sometimes, the resultant distance metric is enough to estimate novelty and risk. However, in many cases more refined postprocessing is required, as described in this section.

Consider two windows which crossed the deviation score threshold and were flagged by the system as alerts. Assume that the older alert has been labeled/annotated by a subject matter expert, then this label can be propagated (with or without similarity-based weighting) and associated with the newer alert. If the most similar alert was deemed a false positive by a subject matter expert, the newer alert could be discounted. Furthermore, we note this logic can be bootstrapped: starting off with a seed of annotated alerts, future alerts can be annotated on the fly, and these generated annotations can help annotate more alerts in an ongoing process.

However, this process, particularly when labels are bootstrapped, poses the risk of propagating labeling errors downstream without the potential for a correcting signal. A scheme which mitigates this risk is that of \emph{majority voting}: rather than propagate the label of the similar

The procedure is as follows:
\begin{itemize}
\item Given a new alert, compare it to all past alerts. Filter all alerts which do not meet a preset similarity threshold $t_{\textit{cutoff}}$.  Subsequently, select the $k$ most similar of the remaining alerts.
\item Aggregate and average all labels (in the scope of this work, simply binarized to be either false or true). The aggregation may or may not be with weighting by similarity. 
\item If the total average vote is above a certain threshold, assign a positive label to the new alert. Otherwise, assign it as false.
\item If not enough votes are available (meaning the alert is novel and not sufficiently similar to preexisting alerts), assign a positive label to the alert, erring on the side of caution in the face of a previously unobserved deviation from normal.
\end{itemize}

A pseudocode summary of the majority voting algorithm is presented in Algorithm \ref{alg2}.

\begin{algorithm}
\caption{Majority Voting Alert Label Propagation}
\label{alg2}
\begin{algorithmic}[1]
 \REQUIRE new alert window $w_d$, past windows $W$, similarity threshold $t_{cutoff}$, voting set size $k$, anomaly threshold $T_{anom}$
 \STATE initialize $S = \{\}$ \COMMENT{voting candidates set}
 \FOR{$w \in W$}
    \IF{$\textit{d}(w, w_d) < t_{cutoff}$}
        \STATE \textbf{continue}
    \ENDIF
    \STATE add $w$ to $S$
 \ENDFOR
 \STATE sort $S$ by $\textit{d}(\bullet, w_d)$
 \STATE define voting set $S_k \gets k$ closest windows in $S$
 \STATE let $A(w)$ be its anomaly score; Compute mean score $a_{S_k} = \sum_{w_k \in S_k} A(w_k)$
 \IF{$a_{S_k} \geq T_{anom}$}
    \STATE set $A(w_d) = 1$ \COMMENT{anomalous}
 \ELSE
    \STATE set $A(w_d) = 0$ \COMMENT{normative}
 \ENDIF
\end{algorithmic}
\end{algorithm}

\begin{figure}[htb]
\centering
\includegraphics[width=\columnwidth]{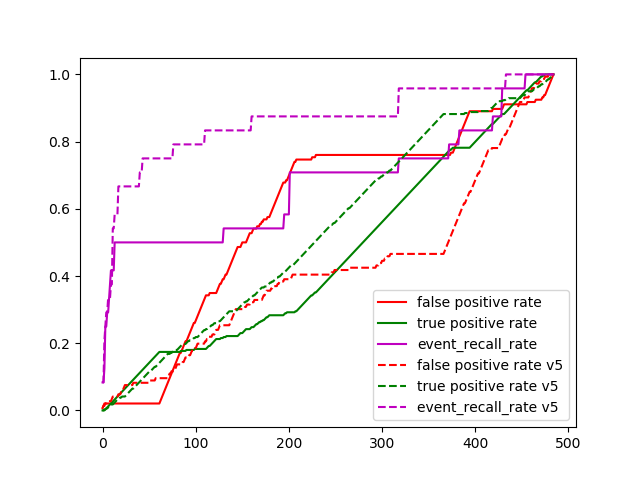}
\caption{
Event recall, alert true and false positive rates with and without using voting for a specific configuration of the framework. The dotted lines reflect a voting pool of 5 rather than the 3 most similar previous alerts. This figure highlights the impact of the majority voting and label propagation scheme on alerting quality proposed in Section 3.2. False positive and true positive rates apply to individual alerts, whereas event recall refers to events (the distinction between alerts and events is made in Section 1 of the paper
%Event recall, alert recall, false positive rates with and without using voting for a specific configuration of the framework. The dotted lines reflect a voting pool of 5 rather than the 3 most similar previous alerts.
}
\label{fig2}

\end{figure}

This approach can substantially reduce the number of false positives without affecting event recall, as shown in Figure \ref{fig2}. We also note that our proposed approach also serves to alleviate explainability concerns, as it enables the alignment of new alerts with past ones, and the identification of specific features and patterns which are most relevant in a given moment.

\begin{figure}[htb]
\centering
\includegraphics[width=\columnwidth, height=200pt]{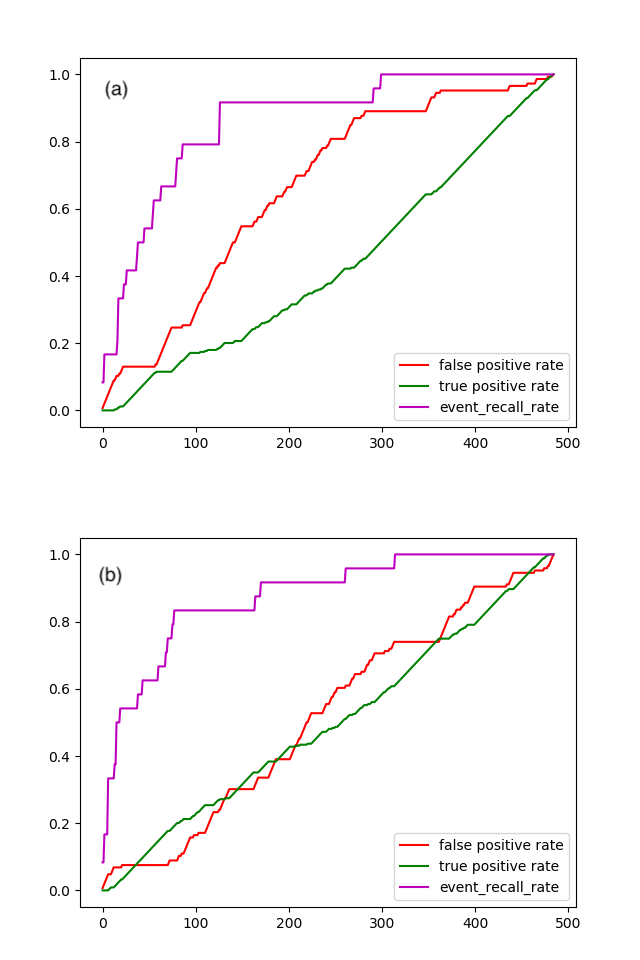}
\caption{(a) Event recall, alert recall, false positive rates with baseline (feature importance cosine distance) similarity measure; (b) Event recall, alert recall, false positive rates with time-deviation based similarity measure }
\label{fig3}

\end{figure}

Note that using a time-series-based similarity measure improves precision without hurting event recall by about 10-15\%, as evidenced in Figures \ref{fig3}(a), \ref{fig3}(b)

% \begin{figure}[htb]
% \centering
% \includegraphics[width=\columnwidth]{Figure27.png}
% \caption{}
% \label{fig4}

% \end{figure}

We discuss in Section \ref{app:opt} how Gaussian Process Optimization can be applied to improve performance in sample.

%\subsection{}

\subsection{Risk Modeling Across Time}
\label{risk_modeling_time}
\begin{figure}[htb]
\centering
\includegraphics[width=250pt, height=150pt]{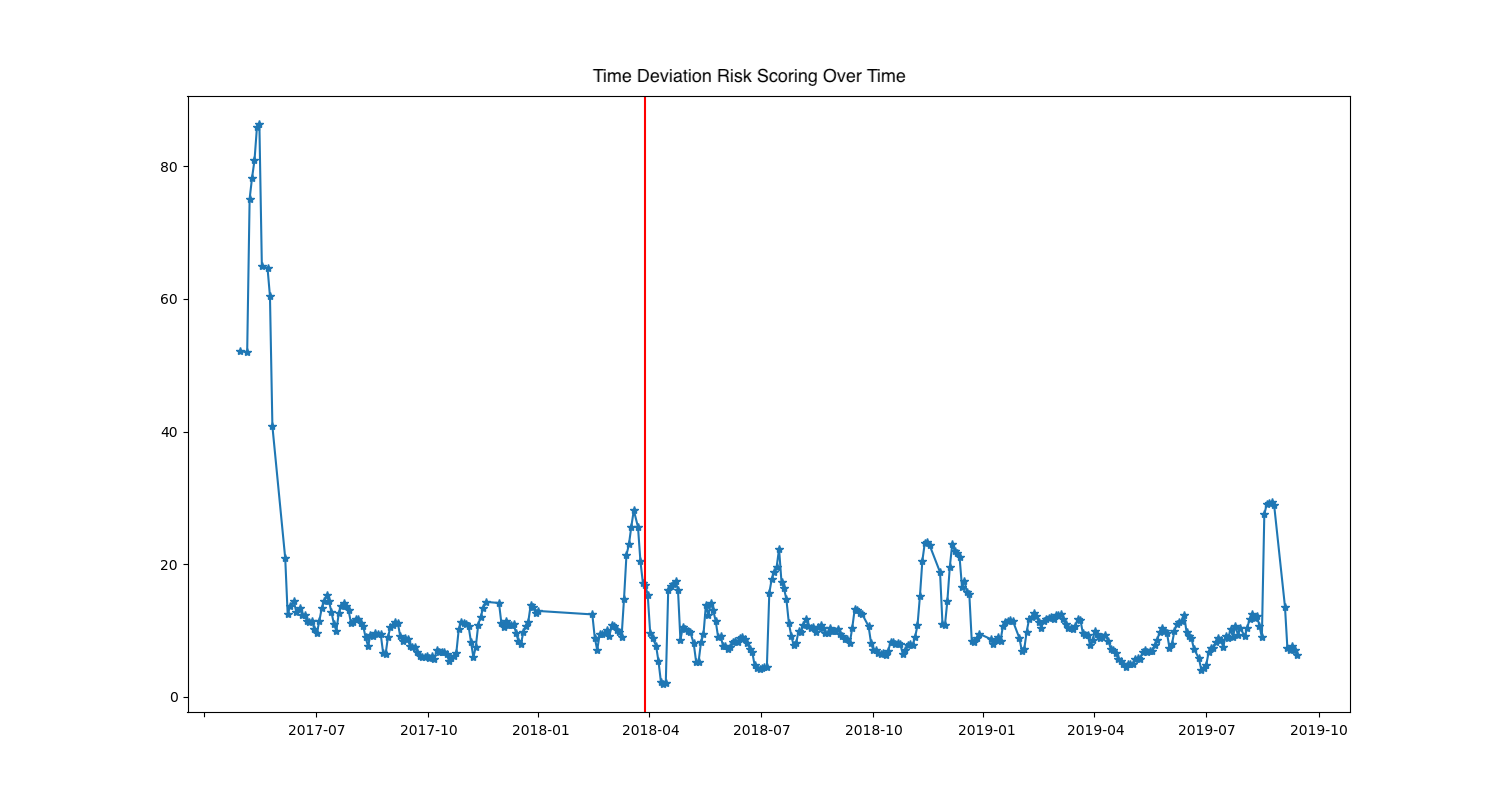}
\caption{
Computation of temporal deviation score before an event starts based on the most informative features prior to an anomalous event. As desired, a marked rise in deviation, representing risk, directly precedes a human-annotated anomalous event. Note that the initially high anomaly score for the first few samples is an artifact resulting from the fact that with very little history, most new windows would appear novel. This time deviation score is functionally analogous to the reconstruction error of an autoencoder in terms of how it can be used for direct alerting.}
%Computation of temporal deviation score before an event starts based on event feature importance.. As desired, a marked rise in risk directly precedes a human-annotated anomalous event. Note that the initially high anomaly score is an artifact resulting from the fact that with very little history, most new windows would appear novel. }
\label{fig5}

\end{figure}

As mentioned earlier in this paper, while the approach discussed in Section \ref{sec:alsim} is useful, the ramifications of the analysis framework described in Section \ref{sec:method} go well beyond merely comparing alerts generated by a different model. Indeed, in this section we expand on how our approach can be used directly as a separate alerting mechanism which differs from and augments the risk output generated by an autoencoder~\cite{zhou2017anomaly}, a commonplace baseline approach to anomaly detection. This application may function as a ``second tier of defense'' against abnormalities missed by the baseline model, and may also be used to suppress false positives even before the alert filtering process described in Section \ref{sec:alsim}.
The core insight is that, as also stated in Section \ref{sec:core}, we can compute a rolling score of ``novelty'' to each new slice in time \emph{relative to all past windows}. In Figure \ref{fig5} we show an example for how this risk scoring approach behaves.

Given the relatively lightweight distance measures we are using, such a comparison, though costlier than only comparing alerts to one another, is still feasible. However, if runtime becomes a bottleneck, an approximation of this score can be done via comparing only parts of windows, utilizing efficient caching to reduce the comparison such that the distance between pairs of obviously different windows does not get fully computed, and, as stated in Section \ref{sec:core}, limiting the time-depth for window comparison (this approach is particularly useful in applications in which drift renders any comparison between the present and the distant past irrelevant). 

To illustrate how our time-series pattern deviation approach to risk scoring is complementary to that of an autoencoder approach, we consider a scatter plot of the two risk indices and cross-reference it with actual human-annotated anomalous events (marked red). As can be seen, for a large proportion of actual anomalous time windows, the time deviation risk score is substantially higher than the reconstruction error based risk score.

%
%\begin{figure}[!htb]
%\centering
%\includegraphics[width=0.8\columnwidth]{Figure10.png}
%\caption{Computation of temporal deviation score before an event starts based on event feature importance, example 2}
%\label{fig6}
%
%\end{figure}
%
%\begin{figure}[!htb]
%\centering
%\includegraphics[width=0.8\columnwidth]{Figure11.png}
%\caption{Computation of temporal deviation score before an event starts based on event feature importance, example 3}
%\label{fig7}
%
%\end{figure}

\begin{figure}[htb]
\centering
\includegraphics[width=\columnwidth, height=180pt]{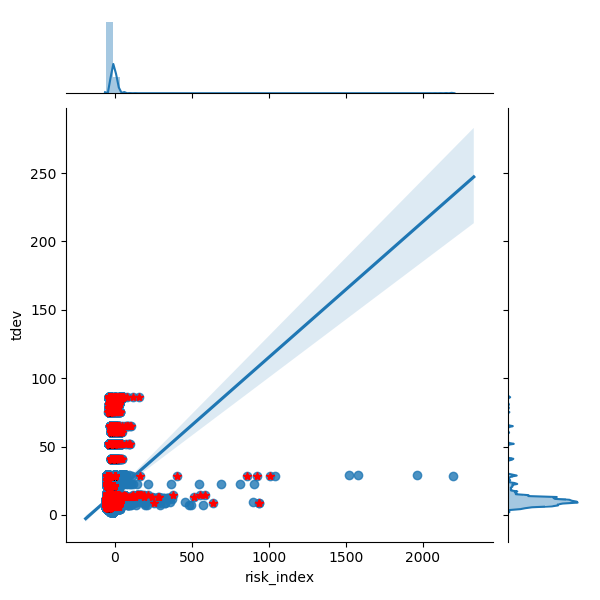}
\caption{ A scatter plot with a fitted linear curve capturing the relationship between the reconstruction-error-based risk score and the temporal deviation risk score, as well as marginal distributions. Each dot represents a sample; the X axis represents the risk index based on autoencoder reconstruction error; the Y axis represents the time deviation score as described in Section 3.1. Dots marked in red were labeled part of actual malfunction events by a human subject matter expert. Note that while the two risk scoring approaches are correlated, the pattern-deviation based approach does a substantially better job of capturing these human-annotated malfunction events.}
%A scatter plot with a fitted linear curve capturing the relationship between the reconstruction-error-based risk score and the temporal deviation risk score. Red highlights the reference event. Note that while the two risk scoring approaches are correlated, the pattern-deviation based approach does a substantially better job of elevating human-annotated anomalous events (marked in red). }
\label{fig122}

\end{figure}

\section{Applicability to Other Datasets}
\label{sec:pub_data}

Focusing on SparkCognition's internal datasets raises the question of whether the performance of our proposed approaches - particularly, the very utility of a contour-based approach for multivariate anomaly detection compared to other approaches (such as autoencoders and TCNs, to name a few). It is also a limitation that due to the size of the SparkCognition's typical datasets, a fair comparison to more complicated techniques driven by more expensively-trained models, such as TCNs, was infeasible. To mitigate these two concerns, we compared our algorithm to a state of the art TCN-based anomaly detection model, using a publicly available dataset (Yahoo's Webscope S5 dataset \cite{thill2017online}). To make the temporal resolution of the data similar to that used when studying the previous results, the data was upsampled by one minute frequency, with values linearly interpolated. Anomalous labels were determined by examining whether any of the underlying features were anomalous in each given time window. We define events in this dataset as contiguous temporal blocks of anomalous behavior. We compare the anomaly scoring of the time-deviation algorithm compared to that of a TCN, similarly to the analysis presented in Section \ref{risk_modeling_time}. As seen in Figure \ref{fig122}, while this dataset is more challenging to this approach, it still does a substantially better job at identifying events as elevated risk events, mostly placing them in the upper 50\% of the risk distribution, implying an elevated risk for an anomaly, compared to the TCN, which fails entirely to capture anomalies in this dataset.

\begin{figure}[htb]
\centering
\includegraphics[width=.8\columnwidth, height=200pt]{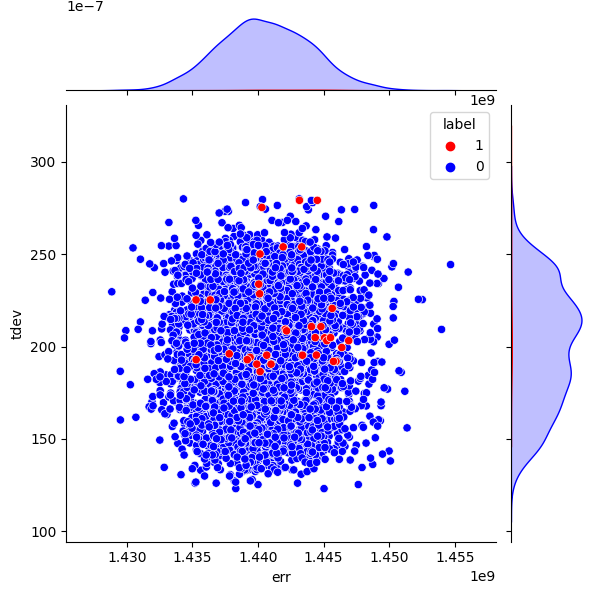}
\caption{A scatter plot capturing the relationship between the TCN-based risk score and the temporal deviation risk score, as well as marginal distributions. Each dot represents a sample. Red dots represent labeled events.}

%Results for optimizing the threshold and number of voting neighbors in the alert label propagation voting scheme. The red asterisk indicates the optimal configuration found.}
\label{fig12}
\end{figure}

\begin{figure}[!htb]
\centering
\includegraphics[width=\columnwidth, height=150pt]{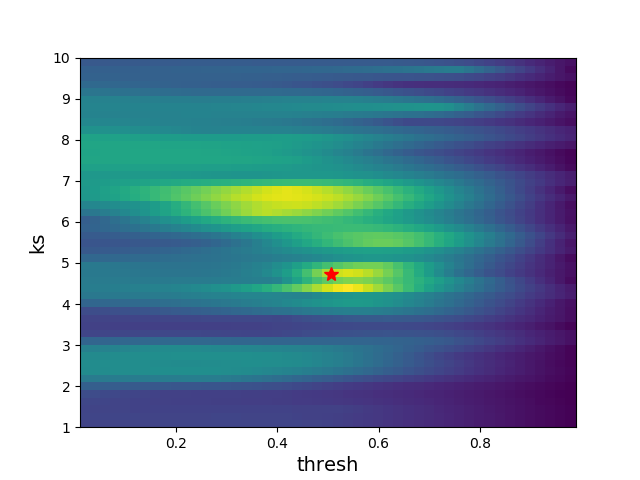}
\caption{Results for optimizing the threshold (what value of aggregated votes constitutes a positive malfunction event label) and number of voting neighbors used to compute this value in our alert label propagation voting scheme, as described in Section 3.2. The value being optimized is the maximal ratio between event recall and false alert rate for a given parameter set, determined at the optimal alert labeling decision criterion for that configuration. The red asterisk indicates the optimal configuration found. It is also evident that the value landscape is fairly smooth, meaning that the selected parameter set has a better chance of robustness.}

%Results for optimizing the threshold and number of voting neighbors in the alert label propagation voting scheme. The red asterisk indicates the optimal configuration found.}
\label{fig11}

\end{figure}

\begin{figure}[!htb]
\centering
\includegraphics[width=220pt, height=150pt]{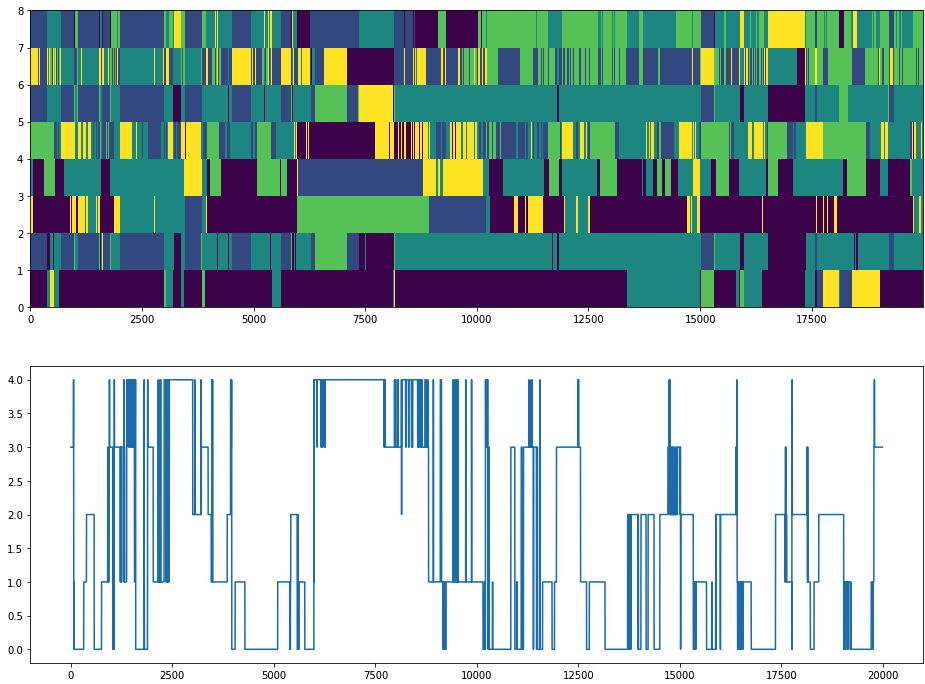}
\caption{Clustering windows into modes based on frequencies of repeated shapelet-type patterns. Top: two-dimensional histogram (color representing frequency) illustrating the frequencies of a small subset of time-series words over time. Bottom: the dominant cluster affiliation resulting from applying k-means on multivariate shapelet frequencies.
}

%Clustering windows into modes based on frequencies of repeated patterns. Top figure displays a two-dimensional histogram to represent the frequencies of a small subset of time-series words. Bottom figure shows the cluster affiliation resulting from applying the canonical k-means algorithm on word frequencies.}
\label{fig10}

 \end{figure}

\section{Optimizing the Framework}
\label{app:opt}

As described in Section \ref{sec:method}, the overall approach presented in this paper relies on multiple degrees of freedom and parameters, such as the temporal window size, the choice of distance measure, the similarity threshold and number of similar windows used in the voting scheme, etc. This fact poses the challenge of tuning the framework and estimating how different parameter sets impact performance. However, as a proof of concept, we show it is possible to \emph{optimize} the framework by defining a target function which captures the ratio between the areas under the event recall and the false alert rate curves. An even simpler shorthand for this ratio, which works well in practice, is the maximum point ratio between the event recall and the false alert for a specific threshold.In other words, as an optimization target, we consider the maximal ratio between event recall and false alert rate for a given parameter set, determined at the optimal alert labeling decision criterion for that configuration. After defining this utility function, we can apply Bayesian optimization~\cite{brochu2010tutorial} to optimize parts of the framework, as illustrated in Figure \ref{fig11}, showing the utility of using different similarity thresholds vs. number of voting similar alerts $k$. Further optimization of other parameter sets is possible. We have found that optimizing small subsets of features at a time rather than all of them at once works better in practice. 

%Finally, in the next section we discuss some additional uses we have found for the proposed approach.
  \section{Additional Applications}

  Beyond its applications to anomaly detection this time series approach holds certain promise in mapping out and categorizing recurring patterns in the data. In that sense, it is a computationally inexpensive way of identifying a multivariate version of shapelets \cite{ye2009time}, which can be leveraged to characterize different modes of operation and characteristic regimes. An example for such analysis is presented in Figure \ref{fig10}. %An example of these multivariate patterns can be shown in a simple 4-feature case is shown in Figure \ref{fig8}. 

\section{Summary and Discussion}

In this paper we present a novel, lightweight approach for anomaly detection in large-scale, real-world applications. Our approached is based on smoothing and segmenting multivariate data into windows, comparing current windows with past ones to determine novelty, and defining the risk as the distance from the most similar pattern observed in the relevant past. This approach can be used by itself or coupled with additional anomaly scoring strategies in order to automatically annotate alerts, filter suspected false alerts, or boost potentially missed anomalous behavior. We show how our framework can also be used to further analyze time-series data, contributing to explainability. Our approach is tunable, optimizable and computationally efficient, and does well to mitigate many of the challenges anomaly detection systems face when handling complex real-world data. There are many potential directions for future work expanding on our proposed method, ranging from subtler ways of hybridizing it with other anomaly detection techniques, utilizing more sophisticated similarity measures and scoring approaches, or using more nuanced ensemble voting for alert filtering. We look forward to the prospect of others studying how applicable our approach is in other real-world datasets and domains.

%TODO:
%* Improve presentation/explanation of figures
%* Finish Summary and Discussion
%* Improve Abstract, Intro
%\bibliographystyle{abbrv}
\bibliography{refs}

\end{document}